\pdfoutput=1

\documentclass[11pt]{article}

\usepackage[preprint]{acl}
\usepackage{amsmath}
\usepackage[utf8]{inputenc}
\usepackage{tcolorbox}
\usepackage{tabularx}
\usepackage{comment}
\usepackage{hyperref}
\usepackage{graphicx} 
\usepackage{multirow} 
\usepackage{float} 

\usepackage{times}
\usepackage{latexsym}
\usepackage{booktabs}
\usepackage[T1]{fontenc}
\usepackage{lipsum}  

\usepackage{amsthm}

\usepackage{array}

\usepackage[utf8]{inputenc}

\usepackage{microtype}

\usepackage{inconsolata}

\usepackage{graphicx}

\definecolor{mLightBrown}{HTML}{EB811B}

\newcommand{\az}[1]{\color{black} #1}

\title{Protoknowledge Shapes Behaviour of LLMs in Downstream Tasks: Memorization and Generalization with Knowledge Graphs}

\author{
\textbf{Federico Ranaldi$^1$}\\
\textbf{Andrea Zugarini$^3$, Leonardo Ranaldi$^{2,1}$}\\
\textbf{Fabio Massimo Zanzotto$^1$} \\
$^1$Human-centric ART, University of Rome Tor Vergata, Italy \\
$^2$School of Informatics, University of Edinburgh, UK \quad
$^3$expert.ai, Italy \\
\\
\small{
\href{mailto:federico.ranaldi@uniroma2.it}{\color{black} \tt federico.ranaldi@uniroma2.it}}
\\
\small{\href{mailto:fabio.massimo.zanzotto@uniroma2.it}{\color{black} \tt fabio.massimo.zanzotto@uniroma2.it}}
}

\begin{document}
\maketitle
\begin{abstract}

We introduce the concept of \emph{protoknowledge} to formalize and measure how sequences of tokens encoding Knowledge Graphs are internalized during pretraining and utilized at inference time by Large Language Models (LLMs).
Indeed, LLMs have demonstrated the ability to memorize vast amounts of token sequences during pretraining, and a central open question is how they leverage this memorization as reusable knowledge through generalization. We then categorize \emph{protoknowledge} into \emph{lexical}, \emph{hierarchical}, and \emph{topological} forms, varying on the type of knowledge that needs to be activated. We measure \emph{protoknowledge} through Knowledge Activation Tasks (KATs), analyzing its general properties such as semantic bias. 
We then investigate the impact of \emph{protoknowledge} on Text-to-SPARQL performance by varying prompting strategies depending on input conditions.
To this end, we adopt a novel analysis framework that assesses whether model predictions align with the successful activation of the relevant \emph{protoknowledge} for each query.
This methodology provides a practical tool to explore Semantic-Level Data Contamination and serves as an effective strategy for Closed-Pretraining models.

\end{abstract}

\section{Introduction}

During pretraining, Large Language Models (LLMs) ingest vast amounts of data, memorizing both factual and structured information \cite{carlini2023quantifyingmemorizationneurallanguage, roberts-etal-2020-much}. Beyond simple memorization, LLMs demonstrate the ability to generalize, transferring learned data from domains encountered during pretraining to solve specific tasks \cite{wang2025generalizationvsmemorizationtracing}. LLMs blend memorized information and acquire the ability to apply it to novel tasks. In the process, this ability can reflect different types of learned information, ranging from recognizing facts to understanding complex structural relations within a domain, depending on the specific task.
\begin{figure}[h]
    \includegraphics[width=\linewidth]{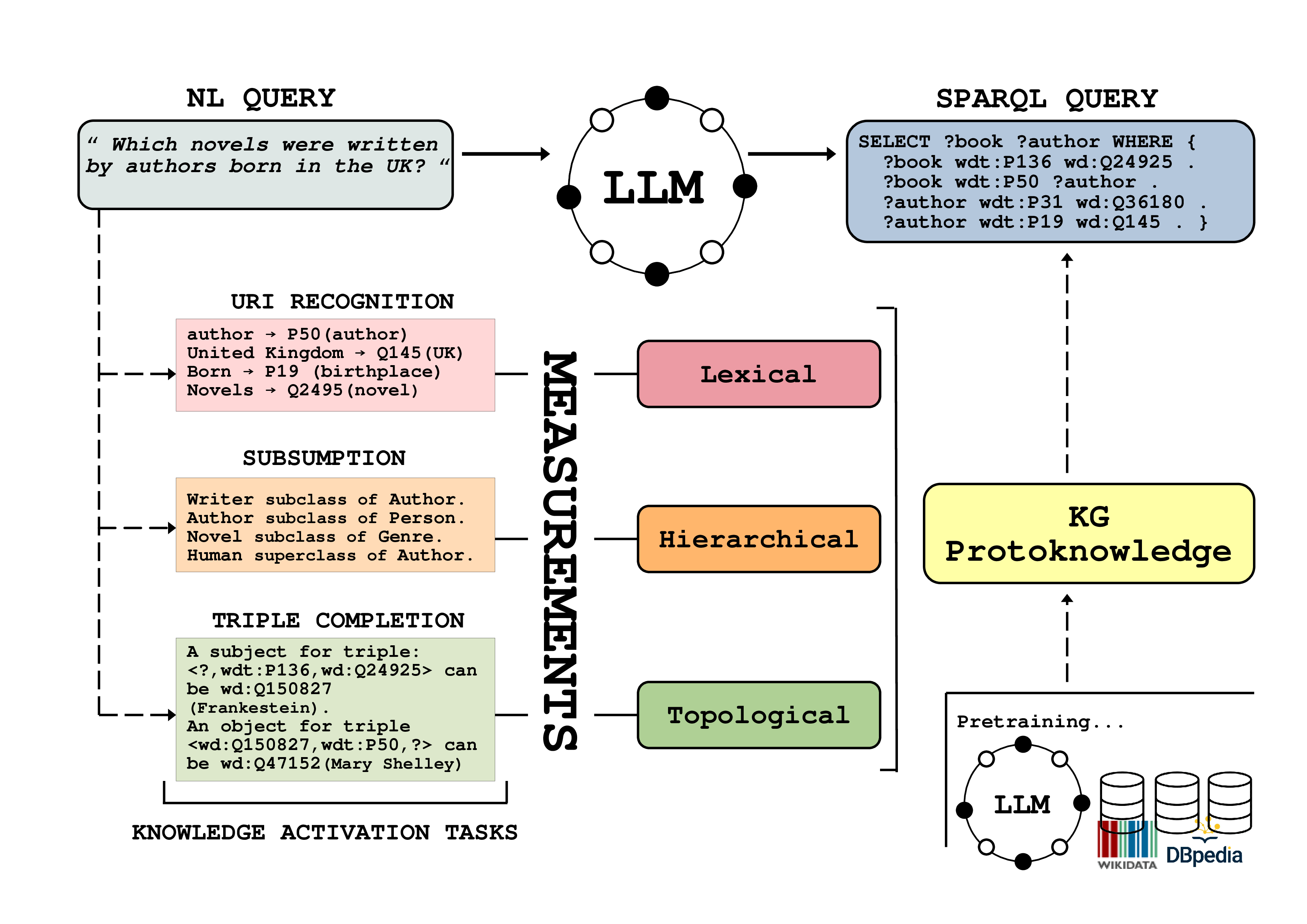}
    \caption{\emph{Protoknowledge} Impact: LLMs acquire three \emph{protoknowledge} forms from Knowledge Graphs, measured via Knowledge Activation Tasks (KATs). The degree of absorption of \emph{protoknowledge} measured by KATs correlates with the Text-To-SPARQL performances of LLMs showing that, when \emph{protoknowledge} is acquired, it is also positively used.}
    \label{fig:sps_evaluation_plot}
    \vspace{-0.3cm}
\end{figure}

\vspace{-0.5cm}

However, the effectiveness of this emerging ability remains limited by the semantic bias inherent in the pretraining data.
This semantic bias confines the capability of models to generalize effectively to new, out-of-distribution domains \cite{elazar2023measuringcausaleffectsdata, wang2025generalizationvsmemorizationtracing}.

Since pretraining often includes structured data ingested as sequences of tokens, such as portions of Knowledge Graphs (KGs), there is a clear opportunity to explore the extent to which LLMs acquire and use this structured information in related task. 

To this end, we introduce the notion of \emph{KG protoknowledge} to study how LLMs exhibit both memorization and generalization over KG content. Recent studies \cite{lo2023exploringreasoningcapabilitylarge, bombieri2025llmsdreamontologies} have examined how LLMs process KGs and leverage KG-based resources without fine-tuning, focusing primarily on memorization. Still, the application of memorized information to generalize and solve KG-related tasks, such as Text-to-SPARQL, remains underexplored.
By studying \emph{KG protoknowledge}, we aim to better understand how models memorize and generalize information in the context of KG-related tasks, particularly by investigating how semantic alignment between the test set and pretraining data influences performance. We hypothesize that \emph{protoknowledge} significantly impacts performance, especially when test set semantics align with the pretraining data.

Building on data contamination \cite{magar-schwartz-2022-data, deng-etal-2024-investigating, dong-etal-2024-generalization} at the semantic level \cite{xu2024benchmarkdatacontaminationlarge}, we design a specific formulation of \emph{protoknowledge} tailored to the type of information that is useful for solving a specific task. We then measure this \emph{protoknowledge} within a benchmark to estimate how much a pretrained model already knows about the underlying semantics of the test set. We hypothesize that a model which has extensively learned semantics related to the benchmark during pretraining is more likely to perform well on that task.
Our contributions are:

\begin{itemize}
\item We formalize \emph{KG protoknowledge} in \emph{lexical}, \emph{hierarchical} and \emph{topological} forms to capture the different levels of abstraction of how Knowledge Graph information is internalized by LLMs during pretraining.

\item We empirically measure \emph{KG protoknowledge} forms by designing \textbf{Knowledge Activation Tasks} and specialized test sets uncovering its alignment with the \textbf{semantic bias} induced by pretraining data as a key property.

\item We investigate the impact of \emph{KG protoknowledge} on Text-to-SPARQL through a per-example analysis that examines how different forms correlate with query generation. This analysis implicitly defines a framework for detecting \textbf{semantic-level data contamination} in Closed-Pretraining Models.

\end{itemize}

\section{Related Works}

The boundary between memorization and generalization in Large Language Models (LLMs) is increasingly relevant, particularly when models are pretrained on large corpora containing structured data such as Knowledge Graphs (KGs). Early research primarily addressed memorization \cite{carlini2023quantifyingmemorizationneurallanguage} and risks of data contamination \cite{magar-schwartz-2022-data,ravaut2024largelanguagemodelscontaminated}. In contrast, more recent studies \cite{wang2025generalizationvsmemorizationtracing} have begun examining how the statistical properties of pretraining corpora shape model outputs, highlighting phenomena like semantic-level data contamination \cite{xu2024benchmarkdatacontaminationlarge}.
In parallel, attention has shifted toward how LLMs internalize KG-based information from pretraining. \citet{bombieri2025llmsdreamontologies} assessed LLMs’ retention of biological ontologies by testing their ability to match entity labels to correct IDs. Similarly, \citet{lo2023exploringreasoningcapabilitylarge} examined GPT models on DBpedia triples, using label prediction tasks that distinguish between exact (Hard Matching) and approximate (Soft Matching) responses. These analyses lead to exploring how such acquired knowledge can be reused in downstream tasks that require implicit reasoning.

A compelling setting to study the impact of KG exposure is the Text-to-SPARQL task. Similar to Text-to-SQL, it requires models to perform structured reasoning, making it well-suited for evaluating whether pretraining on KGs benefits downstream tasks. To the best of our knowledge, Data Contamination in Text-to-SPARQL has not yet been systematically investigated. For models with unknown pretraining (Closed-Pretraining), such analyses must rely on black-box techniques \cite{ravaut2024largelanguagemodelscontaminated}. An effective strategy was proposed by \citet{ranaldi-etal-2024-investigating}, who investigated Spider \cite{yu-etal-2018-spider}, a potentially contaminated Text-to-SQL benchmark, by constructing a structurally equivalent web-independent variant. The observed performance drop in GPTs suggested the original benchmarks were partially memorized at both the verbatim and internal patterns levels. Extending this methodology to Knowledge Graphs is considerably more challenging, given their scale, openness, and entanglement with real-world facts, which renders the construction of contamination-free equivalents infeasible. This motivates alternative approaches, such as probing the models’ internalized information at different levels of abstraction, \emph{KG protoknowledge}, to assess whether their behaviour reflects prior learning on KG content.

\section{Protoknowledge on Knowledge Graphs}

We define \emph{protoknowledge} as the capability of language models to both \emph{memorize} information during pretraining and \emph{functionally reuse} it in downstream tasks, without the need for task-specific supervision or fine-tuning. This capability is particularly salient when solving a task that requires recalling and recombining factual or structural information implicitly acquired from data exposure.

In the context of Knowledge Graphs, \emph{KG protoknowledge} refers to the models' capacity to internalize and leverage information related to entities and properties observed during training.

While previous work has shown that LLMs can memorize portions of KGs at a verbatim level, we expand the analysis to investigate whether models also acquire \emph{abstractions} over KG content that can be reused to support reasoning in KG-related tasks like Text-to-SPARQL.

\subsection{Forms of Protoknowledge}

\emph{KG protoknowledge} can manifest in different forms, depending on the type of information required by the downstream task. Some tasks primarily involve \emph{lexical} mappings, others require structural reasoning, while certain tasks demand both.
Based on the diversity of KG-related tasks, we identify three fundamental types of \emph{protoknowledge}, each reflecting a different abstraction pattern: 
\textbf{\emph{lexical}}, 
\textbf{\emph{hierarchical}}, 
and \textbf{\emph{topological}}.

\paragraph{\emph{Lexical Protoknowledge.}} Refers to a model's ability to recall entities and properties based on natural language surface forms (e.g., labels, aliases). It reflects symbolic anchoring acquired during pretraining. This form of \emph{protoknowledge} is particularly important for tasks that require recovering identifiers from textual inputs, especially when IDs are non-human-readable, as in Wikidata (e.g., mapping the label \texttt{"Moon"} to its URI \texttt{wd:Q405}).

\paragraph{\emph{Hierarchical Protoknowledge.}} Refers to a model’s ability to recognise and reason over taxonomic relations, such as \texttt{subclassOf} or \texttt{instanceOf} hierarchies commonly found in Knowledge Graphs. For example, recognising that \texttt{Politician} is a subclass of \texttt{Person} in DBpedia Ontology. This form of \emph{protoknowledge} can be crucial in tasks like ontology alignment, where understanding the parent-child structure of concepts helps establish semantic correspondences (e.g. in Wikidata \texttt{wd:Journalist} is a direct subclass of \texttt{wd:Writer} and in DBpedia \texttt{dbo:Journalist} is a direct subclass of \texttt{dbo:Person}).

\paragraph{\emph{Topological Protoknowledge.}} Refers to a model's ability to infer and traverse multi-hop relational paths between entities, going beyond direct neighbours. This capability is essential when reasoning over graph structures, e.g., when generating SPARQL queries that require retrieving RDF triples connecting entities via intermediate properties. \emph{Topological} form implicitly requires \emph{lexical} and \emph{hierarchical protoknowledge}, as models must both map labels to URIs and understand their connections inside KGs.

\subsection{Knowledge Activation Tasks}
\label{sec:knowledge_activation_tasks}

To investigate the different forms of \emph{KG protoknowledge}, we rely on a set of evaluation tasks, defined as \textbf{Knowledge Activation Tasks} (KATs). These tasks are designed to selectively activate specific types of \emph{protoknowledge} by constraining the nature of the information and reasoning required to perform them. Although all tasks involve recall and reuse of information acquired during pretraining, they vary substantially in the level of abstraction they expect from models.

Tasks for \emph{lexical protoknowledge} are formulated as label-to-URI translation. The model is presented with a natural language label (or alias) and must output the correct symbolic identifier (typically a URI).

For \emph{hierarchical protoknowledge}, the objective is to test whether the model can infer \emph{hierarchical} relationships between KG items. Models are asked to reconstruct class-subclass relations, either by predicting subclasses from a parent class or by identifying the direct superclass of a given class. This evaluation probes whether the model has internalized implicit relational patterns beyond simple label recognition.

Finally, \emph{topological protoknowledge} tasks examine the model’s ability to reason symbolically over relational paths between entities, requiring inferential steps across graph connections. A typical case is the reconstruction of valid RDF triples, which forms the basis for tasks such as Text-to-SPARQL query generation.

Formalizing KATs for each form provides a scheme for measuring how much \emph{protoknowledge} is encoded in pretrained models and to what extent it is influenced by the semantic bias induced by the pretraining data.
While pretraining corpora of models experimented remain inaccessible, we assume, consistent with prior observations, that their content distribution broadly reflects common web data. Based on this hypothesis, we use our experiments to explore one key limitation of \emph{KG protoknowledge}: its tendency to be strongly influenced by the frequency distribution of KG items, which we interpret as semantic bias.

\section{Measuring Protoknowledge}
\label{sec:fundamental_Protoknowledge}

We define a set of controlled \textbf{Knowledge Activation Tasks} (KATs), designed to selectively activate specific forms of \emph{KG protoknowledge}: \emph{lexical}, \emph{hierarchical}, and \emph{topological}. Each experiment is conducted over dedicated test sets, carefully constructed to isolate the target knowledge type. We evaluate the performance of multiple LLMs, measuring their ability to recall and functionally reuse KG information acquired during pretraining. 

We analyze the correlation between performance and item frequency, supporting the hypothesis that \emph{KG protoknowledge} is influenced and limited by the \textbf{semantic bias of the pretraining data}.

\subsection{Models and Analysis Design}
\label{sec:models}

We evaluate Closed-Pretraining models with undisclosed corpora, making the study of \emph{KG protoknowledge} crucial to indirectly reveal how structured information is internalized and reused by these models. Hence, we use four LLMs\footnote{Model versions in Appendix \ref{app:Models_and_Hyperparameters}}: GPT-4, GPT-3.5-turbo \cite{openai2023gpt4}, Llama-3-8B-Instruct, Llama-3.1-70B and Llama-3-70B-Instruct \cite{grattafiori2024llama3herdmodels}. All experiments adopt greedy decoding to ensure deterministic generation and consistent evaluation.

All experiments are conducted on test sets derived from portions of DBpedia and Wikidata Knowledge Graphs. As a recurrent statistical measure, we define \textbf{popularity} as the number of triples referencing each item, using either the mean or the median (for randomly sampled sets) as a threshold to separate frequent and infrequent items. Additionally, this allows us to analyze how \emph{KG protoknowledge} performance varies across more and less frequent items. Assuming that it reflects the Web bias, we shed light on performance dependence on the distributional properties of the pretraining data. Unless otherwise specified, all analyses on 
\emph{protoknowledge} forms follow the same pipeline with shared configurations and prompting strategies.

\subsection{\emph{Lexical Protoknowledge}}

\paragraph{KATs:} We design the \textbf{URI Recognition Task}, in which the model is prompted with a natural language label (e.g., \texttt{"Moon"}) and asked to predict the corresponding KG URI (Prompt \ref{fig:uri_recognition}). This task assesses the model’s ability to resolve symbolic references from surface forms, a skill especially relevant in settings like Wikidata, where entity identifiers are non-human-readable.

\paragraph{Test Set.} Three test sets of \texttt{(label, Wikidata ID)} pairs are used. The first two are biased toward high-popularity items: one includes the \href{https://www.wikidata.org/wiki/Wikidata:Database_reports/Popular_items}{most common entities}, the other the \href{https://www.wikidata.org/wiki/Wikidata:Database_reports/List_of_properties/Top_100}{most common properties}.
Finally a third set extracted from Wikidata is considered as its distribution content is not biased by high popularity. The extensive description is reported in the Appendix \ref{sec:third_set_analysis}.

\paragraph{Results.} For URI Recognition Task, we report accuracy as the percentage of correct predictions. Results in ~Table \ref{tab:uri_matching_task_combined} show that GPT models significantly outperform Llama models in identifying the most common entities and properties. For instance, GPT-4 achieves an accuracy of 74.35\% on the most frequent entities, whereas Llama-3-70B reaches 35.90\%. 

This disparity highlights GPT models' superior capability in leveraging \emph{lexical protoknowledge} for URI recognition tasks.

\begin{table}[h]
\centering
\small
\setlength{\tabcolsep}{4pt} 
\renewcommand{\arraystretch}{1.1} 
\resizebox{\columnwidth}{!}{ 
\begin{tabular}{@{}lcccc@{}} 
\toprule
\textbf{Index Split} & \textbf{Llama-3-8B} & \textbf{Llama-3-70B} & \textbf{GPT-3.5} & \textbf{GPT-4} \\
\midrule
\multicolumn{5}{c}{\textbf{Most Common Entities}} \\
\midrule
LF (0:161)   & 3.11\% (5/9)  & 19.25\% (31/45)  & 43.46\% (70/94)  & \textbf{48.44\%} (78/107)  \\
MF (-39:200) & 10.25\% (4/9) & 35.90\% (14/45)  & 61.53\% (24/94)  & \textbf{74.35\%} (29/107)  \\
\midrule
\multicolumn{5}{c}{\textbf{Most Common Properties}} \\
\midrule
LF (0:77)    & 2.59\% (2/4)  & 35.06\% (27/38)  & 18.18\% (14/18)  & \textbf{81.81\%} (63/81)  \\
MF (-23:100) & 8.69\% (2/4)  & 47.87\% (11/38)  & 17.40\% (4/18)   & \textbf{78.26\%} (18/81)  \\
\bottomrule
\end{tabular}
}
\caption{URI RECOGNITION accuracy (\%) for Test Set 1 and Test Set 2. LF (Less Frequent) and MFs (More Frequent) subsets are defined based on the average popularity in the dataset. The ratio of correct prediction over the subset of LF or MF is also reported near accuracy.}
\label{tab:uri_matching_task_combined}
\end{table}

\vspace{-0.3cm}

A consistent trend emerges from the results: accuracy tends to be higher for entities and properties that are more frequent/popular. An exception is found in the GPTs' performance on properties, where accuracy remains comparable between more and less popular items. The same trend is also observed in the third set, which was tested only on Llama models for computational reasons (an extended discussion is reported in the Appendix on  Table \ref{tab:uri_matching_task_randomly_selected_entities}).

\subsection{\emph{Hierarchical Protoknowledge}}

\paragraph{KATs:} We focus on taxonomic inference within the DBpedia Ontology and define two subsumption-based tasks:\\
\textbf{Direct Subsumption}: given a class, the model is asked to return its direct subclasses (prompt~\ref{fig:direct_subsumption}).  
\textbf{Inverse Subsumption}: given a class, the model must identify its direct superclass (prompt~\ref{fig:inverse_subsumption}).

While both tasks target \emph{hierarchical} relations, they differ in their cognitive demands. \textbf{Direct Subsumption} may be resolved through shallow pattern recall, as the co-occurrence of class pairs and the \texttt{subclassOf} relation may have been encountered during pretraining. In contrast, \textbf{Inverse Subsumption} requires a higher degree of abstraction: since the concept of Superclass is not represented explicitly for KG items, the model must infer it by generalizing over memorized structural patterns.

\paragraph{Test Set.} For evaluating \emph{hierarchical protoknowledge}, we focus on the top-level classes in the DBpedia Ontology, specifically those that do not have any superclasses (ten root-level classes, e.g. Person, Organization, Place ...). For each of them, we extract their direct subclasses.
In the Direct Subsumption task, the test set consists of pairs formed by a parent class and the full list of its direct subclasses.
In the Inverse Subsumption task, the test set is composed of all subclasses of the root classes, each paired with its corresponding unique superclass. Other details are reported in Appendix Table~\ref{tab:dbpedia_classes_summary}.

\paragraph{Results.}
We report accuracy as the proportion of correct predictions for both the \textit{Subsumption Tasks}. Table~\ref{tab:Direct_Subsumption} on \texttt{Direct Subsumption} shows that GPT-4 consistently outperforms all models, achieving the highest accuracy across both more and less popular classes. GPT-3.5 Turbo follows with strong results on \textit{Person} (80.56\%) and \textit{Organisation} (76.92\%). Llama-3-70B lags behind, prevailing mostly for popular classes, and Llama-3-8B performs poorly on both sets.

\begin{table}[ht]
\centering
\resizebox{\columnwidth}{!}{
\begin{tabular}{c@{\hspace{12pt}}lccccc@{\hspace{12pt}}}
\toprule
\multirow{5}{*}[-12pt]{\small\rotatebox{90}{\textbf{MOST}}} & \textbf{CLASS} & \textbf{Llama-3-8B} & \textbf{Llama-3-70B} & \textbf{GPT-3.5} & \textbf{GPT-4} & \textbf{SUPPORT} \\
\midrule
& Organisation    & 7.69  & 30.77 & 69.23 & \textbf{69.23} & 13 \\
& Place           & 18.18 & 9.10  & 9.09  & \textbf{36.36} & 11 \\
& Work            & 0.00  & 16.67 & 25.00 & \textbf{58.33} & 12 \\
& Person          & 19.44 & 38.89 & 16.66 & \textbf{52.67} & 36 \\
& Species         & 0.00  & 0.00  & 0.00  & 0.00 & 3 \\
\midrule
\midrule
\multirow{6}{*}[-12pt]{\small\rotatebox{90}{\textbf{LEAST}}} & SportFacility           & 25.00  & 0.00  & 0.00  & \textbf{75.00} & 4 \\
& UnitOfWork              & 0.00   & 0.00  & 0.00  & 0.00 & 2 \\
& CelestialBody           & 40.00  & \textbf{80.00} & \textbf{80.00} & \textbf{80.00} & 5 \\
& MeanOfTransportation    & 0.00   & 0.00  & 42.86 & \textbf{85.71} & 7 \\
& ArchitecturalStructure  & 28.57  & \textbf{50.00} & 25.00 & \textbf{50.00} & 4 \\
& Device                  & 25.00  & \textbf{75.00} & 0.00  & 25.00 & 4 \\
\bottomrule
\end{tabular}
}
\caption{\textsc{Direct Subsumption} performance on \textit{Most} and \textit{Least} Frequent Classes.}
\label{tab:Direct_Subsumption}
\end{table}

On \texttt{Inverse Subsumption} results confirm trends about popularity (see Table \ref{tab:Inverse_Subsumption}) that are similar to those of Lexical protoknowledge. GPT-4 leads, exceeding 90\% on four out of five classes and reaching 100\% on Person. GPT-3.5 Turbo maintains good performance for the most popular classes but struggles on the least. Llama-3-70B behaves similarly, predicting the Superclass better for the most popular. Again Llama-3-8B struggles, particularly on rare classes. Notably, GPT-4 maintains high accuracy even for rare categories (e.g., 85.72\% on \textit{MeanOfTransportation}), suggesting its generalization capabilities are less related to KG items' popularity.

\begin{table}[ht]
\centering
\resizebox{\columnwidth}{!}{
\begin{tabular}{c@{\hspace{12pt}}lccccc@{\hspace{12pt}}}
\toprule
\multirow{5}{*}[-12pt]{\small\rotatebox{90}{\textbf{MOST  }}} & \textbf{CLASS} & \textbf{Llama-3-8B} & \textbf{Llama-3-70B} & \textbf{GPT-3.5} & \textbf{GPT-4} & \textbf{SUPPORT} \\
\midrule
& Organisation    & 7.7  & 76.9 & 76.9 & \textbf{92.3} & 13 \\
& Place           & 18.2 & 36.4 & 63.6 & \textbf{90.1} & 11 \\
& Work            & 0.0  & 33.3 & 41.7 & \textbf{100.0} & 12 \\
& Person          & 19.4 & 61.1 & 80.6 & \textbf{94.4} & 36 \\
& Species         & 0.0  & 0.0  &  0.0 & \textbf{33.3} & 3 \\
\midrule 
\midrule
\multirow{6}{*}[-2pt]{\small\rotatebox{90}{\textbf{LEAST}}} & SportFacility           & 0.0  & 0.0  & 0.0  & \textbf{50.0}  & 4 \\
& UnitOfWork              & 0.00  & \textbf{40.0} & 0.0  & 0.0  & 2 \\
& CelestialBody           & 0.0  & 40.0 & 20.0  & \textbf{60.0}  & 5 \\
& MeanOfTransportation    & 0.0  & 42.9 & 0.0  & \textbf{85.7}  & 7 \\
& ArchitecturalStructure  & 0.0  & 25.0 & 25.0  & \textbf{75.00}  & 4 \\
& Device                  & 0.0  & \textbf{75.0} & 50.0  & 70.0  & 4 \\
\bottomrule
\end{tabular}
}
\caption{\textsc{Inverse Subsumption} performance on \textit{Most} and \textit{Least} Popular Classes}
\label{tab:Inverse_Subsumption}
\end{table}

\vspace{-0.5cm}

\subsection{\emph{Topological Protoknowledge}}
\label{sec:topological_Protoknowledge}

\paragraph{KATs:} 
We design Knowledge Activation Tasks (KATs) to assess \emph{topological protoknowledge} via \textit{Subject-Verb-Object} SVO triple completion:
\vspace{-0.1cm}
\begin{itemize}
    \setlength\itemsep{0em}
    \item[] \textbf{SV?:} Given $S$ and $V$, predict $O$ (Prompt \ref{fig:SV_prompt}).
    \item[] \textbf{?VO:} Given $V$ and $O$, predict $S$  (Prompt \ref{fig:VO_prompt}).
\end{itemize}

\paragraph{\emph{Speculative Protoknowledge for SPARQL.}}

We argue that Text-to-SPARQL strongly activates this \emph{protoknowledge} type, since generating correct queries requires retrieving and combining triples. Following \citet{moiseev-etal-2022-skill}, we hypothesize that \emph{protoknowledge} on triples acquired during pretraining enhances this task. We therefore define and measure \emph{topological protoknowledge} speculatively on Text-to-SPARQL benchmarks to assess its impact. However, \emph{topological protoknowledge} can also be studied in any task where triple-centric reasoning is essential.
To study it in the context of Text-to-SPARQL, we introduce a framework that first constructs a dedicated test set and then evaluates the model’s ability to tackle it. We define the \textbf{\emph{Speculative Protoknowledge for SPARQL}} (SPS) as a score:
\begin{equation}
\text{SPS} = \frac{|T_{\text{predicted}} \cap T_{\mathcal{Q}}|}{|T_{\mathcal{Q}}|}
\end{equation}
where $T_{\mathcal{Q}}$ is the set of entity-property pairs extracted from $\mathcal{Q}_{\text{gold}}$ and $T_{\text{predicted}}$ is the set of pairs for which the model correctly inferred a valid triple. Triple validity is verified using SPARQL ASK queries.

\paragraph{Test Set.}
We evaluate SPS on the DBpedia and Wikidata versions of the Text-To-SPARQL Dataset QALD-9plus, extracting all entity-property pairs relevant to $\mathcal{Q}_{\text{gold}}$.

\paragraph{Metrics.}
On generated triples, we distinguish between two levels of correctness: Perfect match, where the predicted URI exactly corresponds to the gold triple, and Soft match, where the predicted URI is related to the correct entity but through a different property. We considered Soft matches as they reflect a form of partially relevant knowledge activation.

\paragraph{Results}
\emph{Topological protoknowledge} is generally stronger on DBpedia than Wikidata, likely due to the human-readable nature of URIs.
In Figure \ref{fig:sps_evaluation_plot}, the GPT models consistently exhibit higher SPS scores than the Llama models in both datasets. Among models, GPT-4 achieves the highest Perfect SPS in all configurations except for the ?VO task on DBpedia, where GPT-3.5 Turbo slightly outperforms it. Comparing these two, GPT-4 shows lower Soft SPS compared to GPT-3.5 Turbo in almost all settings, indicating a more precise activation of \emph{topological protoknowledge}. Within the Llama family, we observe a consistent increase in both Soft and Perfect SPS scores when moving from 8B to the larger 70B, confirming a size-dependent improvement in \emph{topological protoknowledge}. Results on smaller models are displayed in Appendix \ref{tab:sps_results}.

An additional analysis based on popularity (see  Table \ref{tab:popularity_bias_topological}) shows that correctly predicted triples are typically associated with higher popularity of the involved KG items. GPT-4 completed 40 triples (on both tasks) with joint perfect accuracy, of which 27 (67.5\%) exceeded the median popularity; similar trends were observed for GPT-3.5 (19/28), Llama-3-70B (8/12), and Llama-3.1-8B (7/11).

\begin{figure}[h]
    \includegraphics[width=\linewidth]{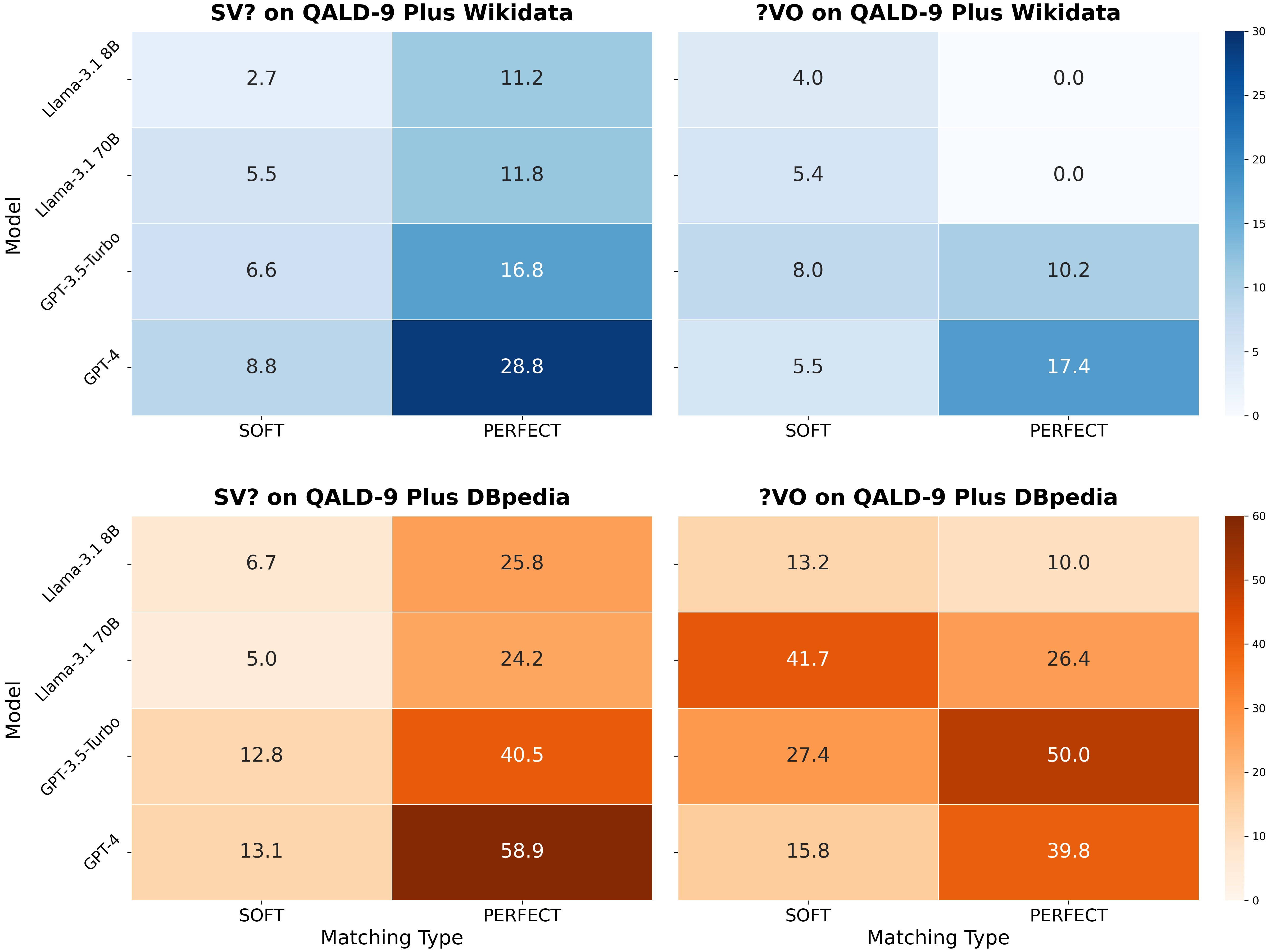}
    \caption{SPS scores for Wikidata and DBpedia.}
    \label{fig:sps_evaluation_plot}
    \vspace{-0.3cm}
\end{figure}

\subsection{General Comment: Semantic Bias and Overconfidence}

Popular entities and properties of knowledge graph (KG) show higher accuracy in measurements of \emph{lexical}, \emph{hierarchical}, and \emph{topological protoknowledge} 
This result 
is robust across different models and tasks, confirming that \emph{protoknowledge} reflects strong dependence on Web content distribution.

Assuming that models' pretraining content aligns with the distributional properties of general-domain KGs like Wikidata and DBpedia, we interpret this as evidence of a persistent \textbf{semantic bias}. Models internalize and reuse KG information more effectively for highly popular items. This contributes to an observed \textbf{overconfidence effect}: performance is stronger when test data aligns with dominant semantic patterns present in the pretraining data.

These findings emphasize the need to account for semantic distribution when designing KG-related evaluation benchmarks. Moreover, performance should be interpreted not only based on the skills elicited by the task itself, but also on how well the test data aligns with the latent frequency priors of the pretraining corpus.

\section{Protoknowledge shapes Text-To-SPARQL}

We analyze how different forms of \emph{KG protoknowledge} influence model performance in Text-to-SPARQL, where the core challenge lies in semantically accurate selection and positioning of KG entities and properties. Using the KGQA framework, we test models under three prompting setups with increasing contextual support, each requiring different levels of implicit reasoning and knowledge activation.

\vspace{-0.2cm}
\paragraph{Experimental Setting.} {\az The baseline prompt (referred to as \texttt{Original}) relies on zero-shot from ~\citet{d2024dynamic}, where for each question, all the entities and relations URIs needed to build the query (and their label) are given as input context. In \texttt{No Label}, URIs are not accompanied by their corresponding labels, thereby requiring the model to recognize them based on its acquired information.
In \texttt{No URI} instead, no additional information is given to support the model.
On KGs like DBpedia, where labels are incorporated in the URI, we limit the experiments to \texttt{Original} and \texttt{No URI} variants.
Prompts are summarized in Table \ref{tab:prompt_configurations}. For more details, see the prompts in Appendix \ref{app:Text2SPARQL_Prompts}. 

\vspace{-0.2cm}
\paragraph{Datasets.} We considered QALD-9~\cite{Unger:2012}, referred to as \textbf{QALD-9 DB}, a KGQA dataset with queries based on DBpedia, and a parallel version based on Wikidata, QALD-9 Plus~\cite{qald9plus} referred to as \textbf{QALD-9 WD}.
\vspace{-0.2cm}
\paragraph{Metrics.} Performance is evaluated using the F1 score, measuring overlap between answers from the target and predicted SPARQL queries. An F1 score of 1 is assigned when both queries return an empty set, and the final score is obtained by averaging all the examples.

\vspace{-0.2cm}
\paragraph{Results.}

Table \ref{tab:comparing_Text2SPARQL_Results_big} , shows a consistent performance degradation across all models from \texttt{Orginal} to \texttt{No URI} according to the richness of contextual information.

On QALD-9 Wikidata, GPT-4 outperforms all other models across all settings and exhibits the smallest drop in \texttt{No Label}, suggesting the influence of \emph{lexical protoknowledge}. GPT-3.5 Turbo, while weaker overall, shows a similarly small drop from \texttt{Original} to \texttt{No Label}. Llama models have a similar trend, with Llama-3.1\_70B performing better in \texttt{No URI}.

On QALD-9 DBpedia, GPT-4 again leads in \texttt{No URI}, while Llamas slightly outperform it in the \texttt{Original} setting, with Llama-3.1\_70B remaining competitive across prompts. GPT-3.5 Turbo performs poorly across all configurations. Full results for smaller models are reported in Table \ref{tab:comparing_Text2SPARQL_Results}.

\begin{table}[h]
    \centering
    \small  
    \renewcommand{\arraystretch}{1.15}  
    \setlength{\tabcolsep}{4pt}  
    \begin{tabular}{llcc}
        \toprule
        \textbf{Model} & \textbf{Approach} & \textbf{QALD-9 WD} & \textbf{QALD-9 DB} \\
        \midrule
        \multirow{2}{*}{Llama-3.1 70B} 
        & \texttt{Original} & 56.34 & 61.45 \\
        & \texttt{No Label} & 47.7 & - \\
        & \texttt{No URI} & 13.41 & 13.94 \\
        \midrule
        \multirow{2}{*}{Llama-3 70B} 
        & \texttt{Original} & 57.88 & 62.79 \\
        & \texttt{No Label} & 48.1 & - \\
        & \texttt{No URI} & 4.20 & 24.87 \\
        \midrule
        \multirow{2}{*}{GPT-4} 
        & \texttt{Original} & 62.74 & 59.32 \\
        & \texttt{No Label} & 58.0 & - \\
        & \texttt{No URI} & 25.14 & 29.09 \\
        \midrule
        \multirow{2}{*}{GPT-3.5-Turbo} 
        & \texttt{Original} & 23.79 & 47.21 \\
        & \texttt{No Label} & 20.7 & - \\
        & \texttt{No URI} & 3.67 & 9.29 \\
        \bottomrule
    \end{tabular}
    \caption{Text-To-SPARQL Performance.}
    \vspace{-0.25cm}
    \label{tab:comparing_Text2SPARQL_Results_big}
\end{table}

\begin{figure*}
    \centering
    \includegraphics[scale=0.2]{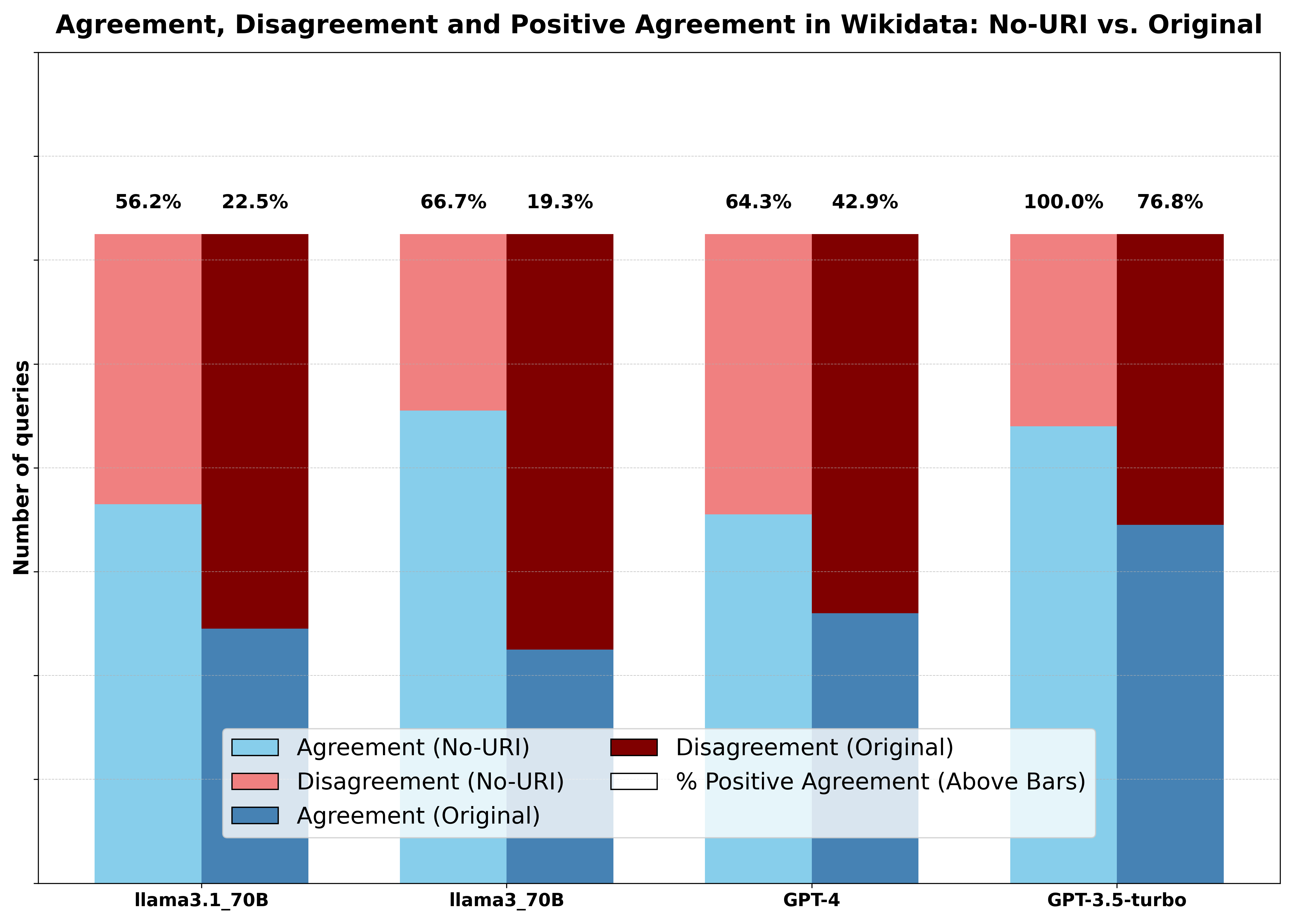}
\includegraphics[scale=0.2]{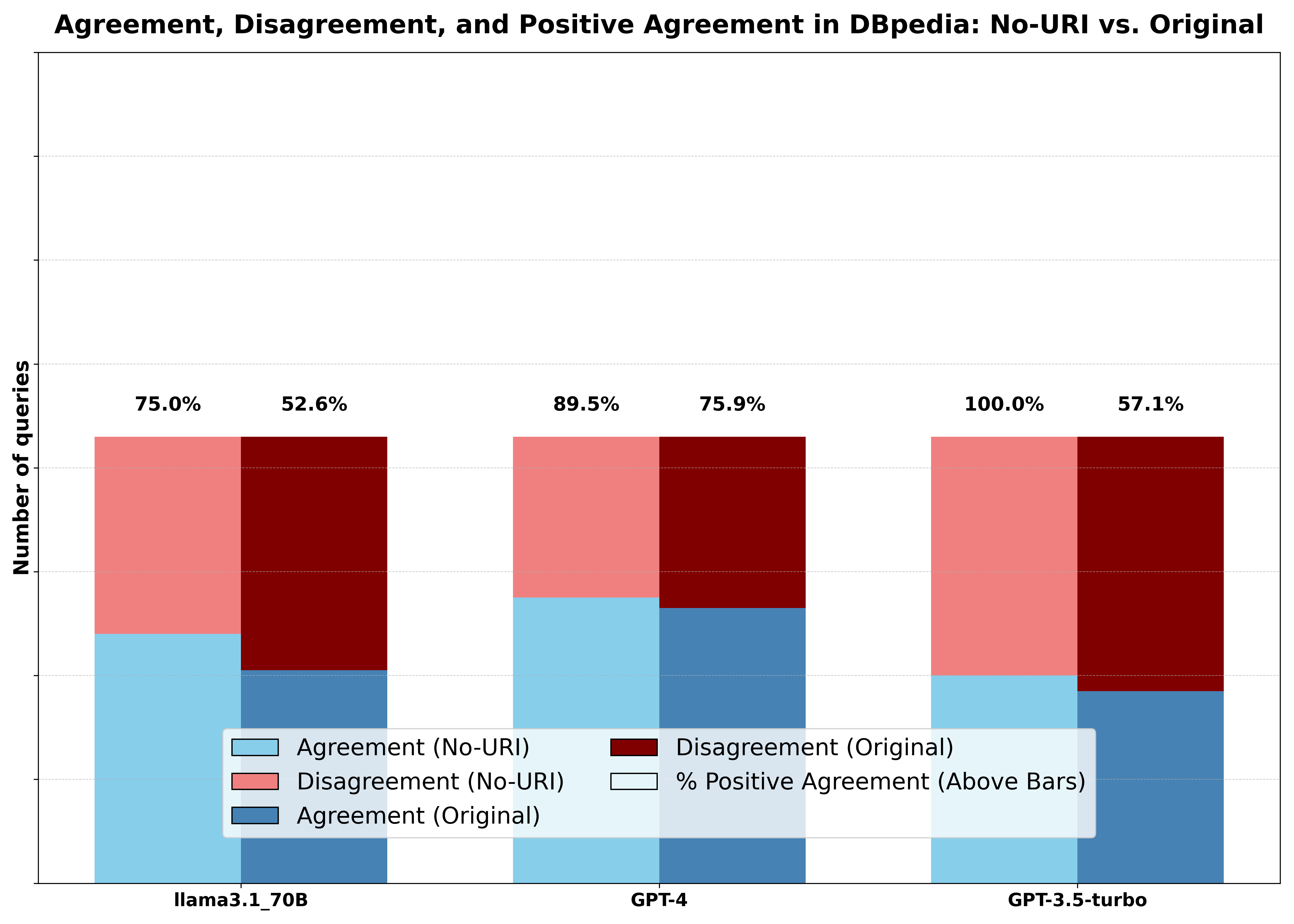}   \caption{NO-URI vs Original Agreement and Disagreement. Above bar is reported Positive Agreement Ratio \%.}
    \label{fig:positive_agreement_WD_DB}
    \vspace{-0.4cm}
\end{figure*}

\vspace{-0.3cm}

\subsection{Impact of \emph{Protoknowledge}}

\paragraph{Framework}
To study the influence of \emph{KG protoknowledge} on \textbf{Text-to-SPARQL} performance, we assess for each query whether the relevant \emph{protoknowledge} type on its content is correctly activated. The general assumption is that a correct SPARQL query implies successful activation of \emph{protoknowledge}. The reverse is not always true: for \emph{lexical} and \emph{hierarchical protoknowledge}, correct \emph{protoknowledge} does not guarantee full query success. Conversely, for \emph{topological protoknowledge}, being more exhaustive and measured speculatively for the task, we hypotesize also the reverse. Hence, we measure on all forms \textbf{Positive Agreement} (PA) as the cases where correct SPARQL generation is accompanied by correct \emph{protoknowledge} activation on its content. Additionally, for \emph{topological} form we measure \textbf{Agreement} the cases where SPARQL query generation and \emph{protoknowledge} activation are both correct or incorrect (complementary cases are measured with \textbf{Disagreement}). A brief description is reported in Appendix Sec.~\ref{sec:framework_brief}.

\vspace{-0.2cm}

\paragraph{Topological}
As observed in the results (Table~\ref{tab:comparing_Text2SPARQL_Results_big}), all LLMs struggle when no additional KG information is provided.
Performance in \texttt{No URI} strongly correlates with SPS scores: models tend to fail in Text-to-SPARQL when they also fail the corresponding triple prediction. Thus, without URI cues, successful query generation relies heavily on the model’s \emph{topological protoknowledge} of the entities and properties involved.

To capture this, we compare query-by-query the model’s SPS scores (Figure~\ref{fig:sps_evaluation_plot}) and Text-to-SPARQL performance in both \texttt{Original} and \texttt{No URI} conditions.

Figure~\ref{fig:positive_agreement_WD_DB} shows that in \texttt{No URI}, Agreement consistently exceeds Disagreement, confirming high reliance on \emph{protoknowledge} when context is absent. Conversely, in \texttt{Original}, Disagreement dominates slightly, indicating a shift towards contextual information. Nonetheless, \emph{topological protoknowledge} remains relevant even with full input. Notably, GPT-3.5 achieves 100\% Positive Agreement in \texttt{No URI}, fully relying on \emph{protoknowledge} for correct queries without additional context.

We also examine whether something analogous holds across KGs. Comparing Positive Agreement between Wikidata and DBpedia (Figure~\ref{fig:no_uri_DC_correlation} and Figure~\ref{fig:original_DC_correlation}), we observe lower values on Wikidata in most models. However, in the \texttt{No URI} setting, the difference in Positive Agreement between the two KGs is minimal, with GPT-3.5 even achieving 100\% Positive Agreement on both. The exception is Llama-3\_70B, where Positive Agreement is based on only three examples (this ratio decreases further in the \texttt{Original} approach).

In \texttt{Original}, the gap between DBpedia and Wikidata widens again, with DBpedia consistently higher. GPT-4, however, maintains high Positive Agreement on Wikidata, reflecting its strong performance on this KG.

\vspace{-0.2cm}

\paragraph{Lexical}

We analyze the impact of \emph{lexical protoknowledge} in the \texttt{No Label} setting, where non-human-readable URIs are provided in the input prompt (considering QALD-9 WD experiments), and no explicit mapping to label is given. In this scenario, the model must implicitly solve the link between the natural language query containing surface forms or approximate label mentions and the correct KG identifiers. A visual description is reported in Appendix~\ref{fig:lexical_protoknowledge_impact_image}.

We report for a subset of models (Tab~\ref{tab:lexical_positive_agreement}) the Positive Agreement as the ratio of correctly generated queries for which \emph{lexical} Protoknowledge was also measured successfully (see example in Fig.~\ref{fig:lexical_protoknowledge_impact_image}). The analysis excludes items below the 50th percentile of popularity (see Tab~\ref{tab:wikidata_items_stats} in Appendix). It's clearly observable that in most cases, the Positive Agreement ratio is very high, achieving 100\% for GPT-3.5 Turbo.

\begin{table}[ht]
\vspace{-0.1cm}
\centering
\resizebox{\columnwidth}{!}{
\begin{tabular}{lcccc}
\toprule
\textbf{Model} & \textbf{Llama-3\_70B} & \textbf{Llama-3.1\_70B} & \textbf{GPT-3.5 Turbo} & \textbf{GPT-4} \\
\midrule
\textbf{PA ratio}     & 28/40  & 15/27 & 13/13 & 41/50 \\
\bottomrule
\end{tabular}
}
\caption{Positive Agreement with \emph{lexical protoknowledge} on QALD-9 WD}
\label{tab:lexical_positive_agreement}
\vspace{0.3cm}
\centering
\resizebox{\columnwidth}{!}{
\begin{tabular}{lcccc}
\toprule
\textbf{Model} & \textbf{Llama-3\_70B} & \textbf{Llama-3.1\_70B} & \textbf{GPT-3.5 Turbo} & \textbf{GPT-4} \\
\midrule
\textbf{PA ratio}     & 32/40  & 35/40 & 22/31 & 40/43 \\
\bottomrule
\end{tabular}
}
\caption{Positive Agreement with \emph{hierarchical protoknowledge} on QALD-9 DB}
\label{tab:hierarchica_positive_agreement}
\vspace{-0.3cm}
\end{table}

\vspace{-0.4cm}

\paragraph{Hierarchical}

We analyze \emph{hierarchical protoknowledge} on the DBpedia version QALD-9 DB, extending the evaluation beyond DBpedia Ontology classes to include other \emph{hierarchical} relations such as \texttt{rdf:type} and \texttt{rdf:subpropertyOf}. Here we analyze the Positive Agreement as the ratio of correctly generated queries for which \emph{hierarchical protoknowledge} was successful.

Results in Tab~\ref{tab:hierarchica_positive_agreement} show the presence of \emph{hierarchical protoknowledge} in successful completions. GPT-4 achieves the highest alignment with a score of 40 of 43 correct queries, followed by Llama-3.1\_70B (35/40), Llama-3\_70B (32/39) and GPT-3.5 Turbo (22/31). These findings suggest that correct query generation strongly correlate with models' ability to internalize and reuse \emph{hierarchical} relations acquired from pretraining.

\vspace{-0.2cm}
\section{Conclusions}

We formalize three core forms of \emph{protoknowledge} within the context of Knowledge Graphs: \emph{lexical}, \emph{hierarchical}, and \emph{topological}. By designing targeted Knowledge Activation Tasks and constructing dedicated test sets, we show that \emph{KG protoknowledge} strongly aligns with web-scale semantic bias, which we hypothesize to be reflected in the pretraining of LLMs.
We further propose a novel framework that correlates Text-to-SPARQL performance with specific forms of \emph{KG protoknowledge}. Our findings indicate that \emph{protoknowledge} strongly influences models' behaviour in this task. Particularly, \emph{topological} proves the most predictive of Text-to-SPARQL success, given the nature of the structured triple information it captures. Yet, the other forms of \emph{protoknowledge} remain valuable. Indeed, \emph{lexical protoknowledge}, is helpful in tackling the \texttt{No Label} setting, confirming its relevance in tasks requiring URI recognition.

Our study highlights the importance of structured knowledge internalization in LLMs and provides a methodology for assessing semantic-level data contamination, with suited applicability to Closed-Pretraining models.

\section*{Limitations}

Our analysis of \emph{KG protoknowledge} is restricted to DBpedia and Wikidata, limiting insights into other Knowledge Graphs (e.g., Freebase, YAGO). Additionally, a deeper comparison between DBpedia and Wikidata would enhance understanding of how \emph{KG protoknowledge} varies across different KG structures.

Expanding to more Text-To-SPARQL benchmarks beyond QALD-9 Plus (e.g., LC-QuAD, GrailQA) would improve the generalizability of our research.

\emph{Topological protoknowledge} is measured by applying triple completion tasks (SV? and ?VO) lacking the S?O task. This omission was intentional, as properties like \texttt{wd:P31} ("instance of" in Wikidata) are frequently used in KGQA benchmkarks, making the evaluation less effective and unfairly favouring this task over others in the SPS metric.

\emph{Lexical Protoknowledge} was specifically analyzed in Wikidata to highlight the challenge of memorizing non-human-readable URIs. \emph{Hierachical Protoknowledge} was examined systematically within the DBpedia Ontology subgraph which features a well-defined \emph{hierarchical} structure and in a small way in other KG-portions extracted from QALD-9 DB. Both analyses deserve expansion to more samples and other KGs based on specific domains, in order to see if \emph{KG protoknowledge} is still present.

Finally, only GPT-4, GPT-3.5 Turbo, and Llama-3.x were tested. Evaluating newer LLMs and retrieval-augmented models would refine our understanding of \emph{KG protoknowledge} in evolving architectures.

\bibliography{custom}

\newpage

\clearpage

\appendix

\section{Knowledge Activation Tasks Prompts}
\vspace{0.2cm}
\begin{center}
\begin{minipage}{0.48\textwidth}
    \begin{tcolorbox}[title=Task: Guess the URI by Label]
        \scriptsize
        Return the label of the entity with the following Wikidata ID: \texttt{{uc[0]}}. \\
        \textbf{Important:} Answer only a string according to your knowledge of Wikidata. \\
        \textbf{Important:} The string must satisfy the triple: \texttt{<http://www.wikidata.org/*/{uc[0]}, rdfs:label, [YOUR ANSWER STRING]>}. \\
        \textbf{Important:} Do not answer other stuff apart from the English label.
    \end{tcolorbox}
    \captionof{figure}{URI Recognition Task}
    \label{fig:uri_recognition}
\end{minipage}
\hfill
\vspace{1.6cm}
\begin{minipage}{0.48\textwidth}
    \begin{tcolorbox}[title=Task: Inverse Subsumption]
        \scriptsize
        Using your knowledge of DBpedia ontology, provide the URI of the superclass for the resource \texttt{[SUBCLASS]}. \\
        \textbf{Important:} Your response must contain only the URI. \\
        \textbf{Important:} The URI must satisfy the triple: \texttt{<[SUBCLASS], rdfs:subClassOf, [YOUR ANSWER URI]>}.
    \end{tcolorbox}
    \captionof{figure}{Inverse Subsumption Task}
    \label{fig:inverse_subsumption}
\end{minipage}
\hfill
\vspace{0.8cm}
\begin{minipage}{0.48\textwidth}
    \begin{tcolorbox}[title=Task: Direct Subsumption]
        \scriptsize
        Using your knowledge of DBpedia ontology, return only a list of URIs that are direct subclasses of \texttt{CLASS}. \\
        \textbf{Important:} URIs must be connected to \texttt{CLASS} by the property rdfs:subClassOf. \\
        \textbf{Important:} URIs must satisfy the triple: \texttt{<[URI]>,rdfs:subClassOf,CLASS>}.
    \end{tcolorbox}
    \captionof{figure}{Direct Subsumption Task}
    \label{fig:direct_subsumption}
\end{minipage}
\hfill
\vspace{1.7cm}
\begin{minipage}{0.48\textwidth}
    \begin{tcolorbox}[title=Task: SV? Guess the OBJECT]
        \scriptsize
        Considering your knowledge of DBpedia triples, can you fill the masked \texttt{[MASKED\_OBJECT]} with an existing URI inside DBpedia? \\
        \textbf{TRIPLE:} \texttt{\{\{'{S}','{V}',[MASKED\_OBJECT]\}\}} \\
        \textbf{Important:} Answer only the URI! Do not invent URIs.
    \end{tcolorbox}
    \captionof{figure}{SV? Task}
    \label{fig:SV_prompt}
\end{minipage}
\hfill
\vspace{0.8cm}
\begin{minipage}{0.48\textwidth}
    \begin{tcolorbox}[title=Task: ?VO Guess the SUBJECT]
        \scriptsize
        Considering your knowledge of DBpedia triples, can you fill the masked \texttt{[MASKED\_SUBJECT]} with an existing URI inside DBpedia? \\
        \textbf{TRIPLE:} \texttt{\{\{[MASKED\_SUBJECT],'{V}','{O}'\}\}} \\
        \textbf{Important:} Answer only the URI! Do not invent URIs.
    \end{tcolorbox}
    \captionof{figure}{?VO Task}
    \label{fig:VO_prompt}
\end{minipage}
\end{center}

\begin{table}[]

\section{Models and Hyperparameters}
\label{app:Models_and_Hyperparameters}
To get a comprehensive evaluation, we use four different LLMs: GPT-4, GPT-3.5 \cite{openai2023gpt4}, Llama-3.1-8B, Llama-3-8B, Llama3-70b-instruct \cite{grattafiori2024llama3herdmodels}. We use greedy decoding in all experiments to ensure a more deterministic generation process. We set the temperature to 0 and the maximum generation length to 2048. We observed that these settings deliver better and deterministic performance.

\small
\begin{center}
\small
\begin{tabular}{lp{5.5cm}}
\textbf{Model} & \textbf{Version}  \\ 
\hline
GPT-4 & OpenAI API (gpt-4-o)  \\
GPT-3.5-Turbo & OpenAI API (gpt-3.5-turbo)  \\
\hline
Llama-3 8B   &  meta-llama/Meta-Llama-3-8B-Instruct \\
Llama3 70B  &  meta-llama/Meta-Llama-3-70B-Instruct \\
Llama-3.1-8B & meta-llama/Llama-3.1-8B-Instruct\\
Llama-3.1-70B & meta-llama/Llama-3.1-70B-Instruct\\
\hline
\end{tabular}
\end{center}
\caption{Models (huggingface.co). We used the configurations described in Section \ref{sec:models} in the repositories for each model *(access verified on 18 May 2025).}
\end{table}

\newpage
\newpage

\begin{table*}[!b]
\section{Topological protoknowledge: SPS score on parallel Text-To-SPARQL benchmarks}
\centering
\renewcommand{\arraystretch}{1.2} 
\resizebox{\textwidth}{!}{ 
    \begin{tabular}{lllcccccc}
        \hline
        \textbf{Model} & \textbf{Dataset} & \textbf{KG} & \multicolumn{2}{c}{\textbf{S V ?}} & \multicolumn{2}{c}{\textbf{? V O}} & \textbf{Perfect} & \textbf{Loose} \\ \cline{4-7}
        &  &  & \textbf{S} & \textbf{P} & \textbf{S} & \textbf{P} & \textbf{SV+VO} & \textbf{SV+VO} \\ 
        \hline
        \small \textbf{Llama-3 8B} & \small QALD-9 Plus & \small Wikidata & \small 1.00 & \small 0.76 & \small 4.11 & \small 0.3 & \small 2.37 & \small 1.5 \\ \hline
        \small \textbf{Llama-3 70B} & \small QALD-9 Plus & \small Wikidata & \small 5.71 & \small 7.81 & \small 2.11 & \small 8.62 & \small 9.45 & \small 8.27 \\ \hline
        \small \textbf{Llama-3.1 8B} & \small QALD-9 Plus & \small Wikidata & \small 2.71 & \small 11.25 & \small 4.00 & \small 0.00 & \small 7.66 & \small 3.75 \\ \hline
        \small \textbf{Llama-3.1 70B} & \small QALD-9 Plus & \small Wikidata & \small 5.55 & \small 11.77 & \small 5.37 & \small 0.00 & \small 14.52 & \small 13.53 \\ \hline
        \small \textbf{GPT-3.5-Turbo} & \small QALD-9 Plus & \small Wikidata & \small 6.56 & \small 16.77 & \small \textbf{8.01} & \small 10.19 & \small 19.90 & \small 21.05 \\ \hline
        \small \textbf{GPT-4} & \small QALD-9 Plus & \small Wikidata & \small \textbf{8.76} & \small \textbf{28.77} & \small 5.47 & \small \textbf{17.37} & \small \textbf{27.29} & \small \textbf{30.07} \\ \hline
        \hline
        \small \textbf{Llama-3 8B} & \small QALD-9 Plus & \small DBpedia & \small \textbf{20.00} & \small 12.50 & \small 30.90 & \small 10.00 & \small 34.41 & \small 39.39 \\ \hline
        \small \textbf{Llama-3 70B} & \small QALD-9 Plus & \small DBpedia & \small 13.33 & \small 39.16 & \small \textbf{43.48} & \small 25.45 & \small 58.50 & \small 62.62 \\ \hline
        \small \textbf{Llama-3.1 8B} & \small QALD-9 Plus & \small DBpedia & \small 6.67 & \small 25.83 & \small 13.18 & \small 10.00 & \small 25.84 & \small 29.29 \\ \hline
        \small \textbf{Llama-3.1 70B} & \small QALD-9 Plus & \small DBpedia & \small 5.00 & \small 24.17 & \small 41.66 & \small 26.36 & \small 47.79 & \small 51.52 \\ \hline
        \small \textbf{GPT-3.5-Turbo} & \small QALD-9 Plus & \small DBpedia & \small 12.77 & \small 40.55 & \small 27.42 & \small \textbf{50.00} & \small \textbf{62.54} & \small \textbf{66.67} \\ \hline
        \small \textbf{GPT-4} & \small QALD-9 Plus & \small DBpedia & \small 13.05 & \small \textbf{58.88} & \small 15.75 & \small 39.84 & \small 58.96 & \small 63.15 \\ \hline
    \end{tabular}
}
\caption{Results overview on triple completion tasks performed on Llamas and GPTs. The tasks are conducted on two versions of QALD-9 Plus, based on Wikidata and DBpedia. P represents "Perfect" satisfaction of the triple, while S represents "Soft" satisfaction of the triple.
"Perfect SV+VO" reports the joint evaluation of the triples contingent to a predicted query that are satisfied in a "Perfect" manner.
"Loose SV+VO" reports the joint evaluation of the triples contingent to a predicted query that are satisfied either "Perfectly" or in a "Soft" manner.}
\label{tab:sps_results}
\end{table*}

\newpage
\newpage

\newpage

\clearpage
\newpage

\section{Semantic Bias Analysis}

\subsection{URI recognition task evaluated on unbiased random set}
\label{sec:third_set_analysis}
The third set, was built from the Freebase-Wikidata-Mapping dataset (1M+ triples linking Freebase IDs, Wikidata IDs, and \texttt{rdfs:label}), is unsorted by popularity and thus includes both popular or not entities. We randomly sample 1K \texttt{(Wikidata ID, label)} pairs to reduce inductive bias and test \emph{lexical protoknowledge} over a broader distribution.
The first two test sets are inherently biased: they contain the most popular Wikidata items, resulting in very high average frequencies (800K for entities, 2M for properties). In contrast, the third test, containing 1k samples extracted randomly, presents a much lower popularity (50K average against the 800k of the first set).  For computational reasons, this set was evaluated only on Llama models. Even if accuracies scaled negatively, the trend persists: models perform better on more frequent entities.
\begin{table}[ht]
\centering
\footnotesize 
\begin{tabular}{@{}l p{1cm} rr@{}} 
\toprule
\textbf{Criterion} & \textbf{Subset} & \textbf{Llama-3-8B} & \textbf{Llama-3-70B} \\
\midrule
Med.-based & 0:500 & 0.0\% (0/3) & 0.00\% (0/24) \\
Med.-based & 500:1000 & 0.6\% (3/3) & 4.8\% (24/24) \\
Mean-based & 0:912 & 0.21\%(1/3) & 1.35\% (11/24) \\
Mean-based & 912:1000 & 2.29\% (2/3) & 14.94\% (13/24) \\
\bottomrule
\end{tabular}
\caption{URI Recognition Task Accuracy. Two splitting criteria were considered: median-based and mean based. Although the second method of splitting leads to two subsets of incomparable size, both models tend to retrieve the IDs of entities whose popularity is above the avg. The ratio of correct prediction over the subset of LF or MF is also reported near accuracy.}
\label{tab:uri_matching_task_randomly_selected_entities}
\end{table}

\begin{table}[h]
\centering
\small
\begin{tabular}{lcc}
\toprule
\textbf{Model} & \textbf{Above Median} \\
\midrule
GPT-4         & 27/40 \\
GPT-3.5 Turbo & 19/28 \\
LLaMA-3 70B   & 8/12 \\
LLaMA-3.1 70B & 7/11 \\
\bottomrule
\end{tabular}
\caption{Percentage of correctly predicted triples on both triple completion tasks (Perfect Accuracy on Joint $>$ 0.5) whose average popularity exceeds the dataset mean (12.8M) and median (1.9M). Under Above Median is reported the number of triples with an high average popularity (computed on its items) over all correctly completed triples.}
\label{tab:popularity_bias_topological}
\end{table}

\newpage

\section{Statistics on KG items}

\begin{table}[h!]
\centering
\scriptsize 
\renewcommand{\arraystretch}{1.2} 
\begin{tabular}{p{2.3cm} p{3cm} r}
    \toprule
    \textbf{Class} & \textbf{Example Subclasses} & \textbf{Occurrences} \\
    \midrule
    Person & Astronaut, Politician, ... & 3,415,598 \\
    Species & Archaea, Eukaryote, ... & 3,989,728 \\
    Place & WineRegion,Park ... & 1,526,870 \\
    Work & Database, MusicalWork, ... & 1,439,666 \\
    Organisation & NonProfitOrganisation,, ... & 869,554 \\
    Device & Battery, Engine, ... & 72,356 \\
    MeanOfTransportation & SpaceShuttle, Train, ... & 34,768 \\
    CelestialBody & Constellation, Planet, ... & 30,130 \\
    Arch. Structure & Tunnel, Tower, ... & 13,636 \\
    UnitOfWork & Project, Case, ... & 7,100 \\
    SportFacility & CricketGround, RaceTrack, ... & 5,792 \\
    \bottomrule
\end{tabular}
\caption{Popularity of subclasses for top-level DBpedia classes. DBpedia Ontology was choosen for its well-formed hierarchical structure lacking inconsistencies that are common in other KGs.}
\label{tab:dbpedia_classes_summary}
\end{table}

\begin{table}[ht]
\centering
\small
\begin{tabular}{lccc}
\toprule
\textbf{ID} & \textbf{Occurrence} & \textbf{Popularity} & \textbf{Label} \\
\midrule
\textbf{Entities}    &   &    &  \\
Q55        & 2 & 1,058,631  & Netherlands \\
Q82955     & 1 & 1,748,496  & politician \\
Q34        & 1 & 1,777,306  & Sweden \\
Q148       & 2 & 2,340,375  & China \\
Q145       & 1 & 3,181,387  & UK \\
Q183       & 5 & 3,556,517  & Germany \\
Q6581072   & 3 & 5,379,193  & female \\
Q5         & 2 & 5,586,977  & human \\
Q30        & 2 & 7,026,527  & USA \\
Q2         & 1 & 11,862,402 & Earth \\
\textbf{Properties}    &   &    &  \\
P27    & 3  & 5,575,274   & citizenship \\
P569   & 2  & 7,105,365   & birthdate \\
P21    & 3  & 10,044,816  & sex or gender \\
P106   & 12 & 11,964,330  & occupation \\
P131   & 4  & 13,853,240  & located in \\
P17    & 17 & 18,663,328  & country \\
P50    & 4  & 34,011,532  & author \\
P1433  & 1  & 45,835,521  & published in \\
P577   & 2  & 49,376,272  & publication date \\
P31    & 40 & 118,388,701 & instance of \\
\bottomrule
\end{tabular}
\caption{Top 10 entities and properties above the 50th percentile in QALD-9 WD with their dataset frequency, popularity, and label. We observe that the 50\% of the 50th percentile corresponds to the global top 100 properties and the 25\% of the 50th percentile corresponds to the global top 200 entities.}
\label{tab:wikidata_items_stats}
\end{table}

\newpage

\clearpage
\newpage

\begin{figure*}[h]
\section{Text2SPARQL Prompts}
\label{app:Text2SPARQL_Prompts}
    \centering
    
    \begin{minipage}{0.48\textwidth}
        \begin{tcolorbox}[title=Approach: Original]
            \small
            You are an expert in SPARQL and \texttt{\{KG\_name\}}.
            
            Your task is to translate natural language questions into precise SPARQL queries that retrieve the desired information from \texttt{\{KG\_name\}}.
            
            \textbf{Guidelines:}
            \begin{enumerate}
                \item Understand the input: Analyze the question and use the provided Entities and Relations to construct the query.
                \item Construct a valid SPARQL query: Use proper syntax and ensure the query retrieves accurate results from \texttt{\{KG\_name\}}.
                \item Format the output: Enclose the SPARQL query within <SPARQL></SPARQL> tags. Do not output anything else.
            \end{enumerate}

            \textbf{Question:} \texttt{\{question\}}

            \textbf{Entities:} \texttt{\{URI\} (\{label\}), ...} 

            \textbf{Relations:} \texttt{\{URI\} (\{label\}), ...} 

            \textbf{Query:}
        \end{tcolorbox}
        \caption{Original Approach}
        \label{fig:original_prompt}
    \end{minipage}
    \hfill
    \begin{minipage}{0.48\textwidth}
        \begin{tcolorbox}[title=Approach: No Label]
            \small
            You are an expert in SPARQL and \texttt{\{KG\_name\}}.
            
            Your task is to translate natural language questions into precise SPARQL queries that retrieve the desired information from \texttt{\{KG\_name\}}.
            
            \textbf{Guidelines:}
            \begin{enumerate}
                \item Understand the input: Analyze the question and use the provided Entities and Relations to construct the query.
                \item Construct a valid SPARQL query: Use proper syntax and ensure the query retrieves accurate results from \texttt{\{KG\_name\}}.
                \item Format the output: Enclose the SPARQL query within <SPARQL></SPARQL> tags. Do not output anything else.
            \end{enumerate}

            \textbf{Question:} \texttt{\{question\}}

            \textbf{Entities:} \texttt{\{URI\}}, ... 

            \textbf{Relations:} \texttt{\{URI\}}, ...

            \textbf{Query:}
        \end{tcolorbox}
        \caption{No Label Approach}
        \label{fig:no_label_prompt}
    \end{minipage}

    \vspace{1cm}

    \begin{minipage}{0.75\textwidth}
        \begin{tcolorbox}[title=Approach: No URI]
            \small
            You are an expert in SPARQL and \texttt{\{KG\_name\}}.
            
            Your task is to translate natural language questions into precise SPARQL queries that retrieve the desired information from \texttt{\{KG\_name\}}.
            
            \textbf{Guidelines:}
            \begin{enumerate}
                \item Understand the input: Analyze the question and use the provided Entities and Relations to construct the query.
                \item Construct a valid SPARQL query: Use proper syntax and ensure the query retrieves accurate results from \texttt{\{KG\_name\}}.
                \item Format the output: Enclose the SPARQL query within <SPARQL></SPARQL> tags. Do not output anything else.
            \end{enumerate}

            \textbf{Question:} \texttt{\{question\}}

            \textbf{Query:}
        \end{tcolorbox}
        \caption{No URI approach}
        \label{fig:no_uri_prompt}
    \end{minipage}

\end{figure*}

\newpage

\section{Text-To-SPARQL Experiments}

\begin{table}[h]
    \subsection{Text-To-SPARQL Prompt Configurations}
    \centering

    \small
        \begin{tabular}{p{3cm}p{4cm}}  
            \toprule
            \textbf{Approach} & \textbf{Description} \\
            \midrule
            \texttt{Original} & explicits URI-label associations \\
            \texttt{No Label} & URIs w/o label associations \\
            \texttt{No URI} & No any additional informations. \\         
            \bottomrule
        \end{tabular}

    \caption{Text-to-SPARQL Prompt Configurations.}
    \label{tab:prompt_configurations}
\end{table}

\begin{table}[h]
    \subsection{Extensive Text-To-SPARQL results}
    \centering
    \small
    \renewcommand{\arraystretch}{1.15}  
    \setlength{\tabcolsep}{4pt} 
    \begin{tabular}{llcc}
        \toprule
        \textbf{Model} & \textbf{Approach} & \textbf{QALD-9 WD} & \textbf{QALD-9 DB} \\
        \midrule
        \multirow{2}{*}{Llama-3.1 70B} 
        & \texttt{Original} & 56.34 & 61.45 \\
        & \texttt{No Label} & 47.7 & - \\
        & \texttt{No URI} & 13.41 & 13.94 \\
        \midrule
        \multirow{2}{*}{Llama-3 70B} 
        & \texttt{Original} & 57.88 & 62.79 \\
        & \texttt{No Label} & 48.1 & - \\
        & \texttt{No URI} & 4.20 & 24.87 \\
        \midrule
        \multirow{2}{*}{Llama-3.1 8B} 
        & \texttt{Original} & 43.60 & 21.24 \\
        & \texttt{No Label} & 35.3 & - \\
        & \texttt{No URI} & 0.00 & 3.37 \\
        \midrule
        \multirow{2}{*}{Llama-3 8B} 
        & \texttt{Original} & 49.69 & 29.34 \\
        & \texttt{No Label} & 40.29 & - \\
        & \texttt{No URI} & 0.00 & 5.93 \\
        \midrule
        \multirow{2}{*}{GPT-4} 
        & \texttt{Original} & 62.74 & 59.32 \\
        & \texttt{No Label} & 58.0 & - \\
        & \texttt{No URI} & 25.14 & 29.09 \\
        \midrule
        \multirow{2}{*}{GPT-3.5-Turbo} 
        & \texttt{Original} & 23.79 & 47.21 \\
        & \texttt{No Label} & 20.7 & - \\
        & \texttt{No URI} & 3.67 & 9.29 \\
        \bottomrule
    \end{tabular}
    \caption{Performance of Llamas and GPTs on QALD-9 -Plus in Wikidata and DBpedia version on the Original and No URI approaches.}
    \label{tab:comparing_Text2SPARQL_Results}
\end{table}

\clearpage

\section{Protoknowledge Analysis Impact}

\subsection{Protoknowledge Analysis Framework in brief}
\label{sec:framework_brief}
Given a form of \emph{KG Protoknowledge} and a Text-to-SPARQL query instance, we perform the following steps:

\begin{enumerate}
    \item Extract a mini test set from the query, containing relevant KG elements.
    \item Evaluate \emph{protoknowledge} on this mini set using Knowledge Activation Tasks (KATs).
    \item If both the \emph{protoknowledge} evaluation and the SPARQL generation are correct, the instance is marked as a \textbf{Positive Agreement}.
\end{enumerate}

\subsection{Example of Framework Application on \emph{lexical} form}
We report an example (Fig. \ref{fig:lexical_protoknowledge_impact_image}) applying the framework for correlating \emph{lexical protoknowledge} and Text-To-SPARQL in \texttt{No Label} approach.

\begin{figure}[h]
    \includegraphics[width=\linewidth]{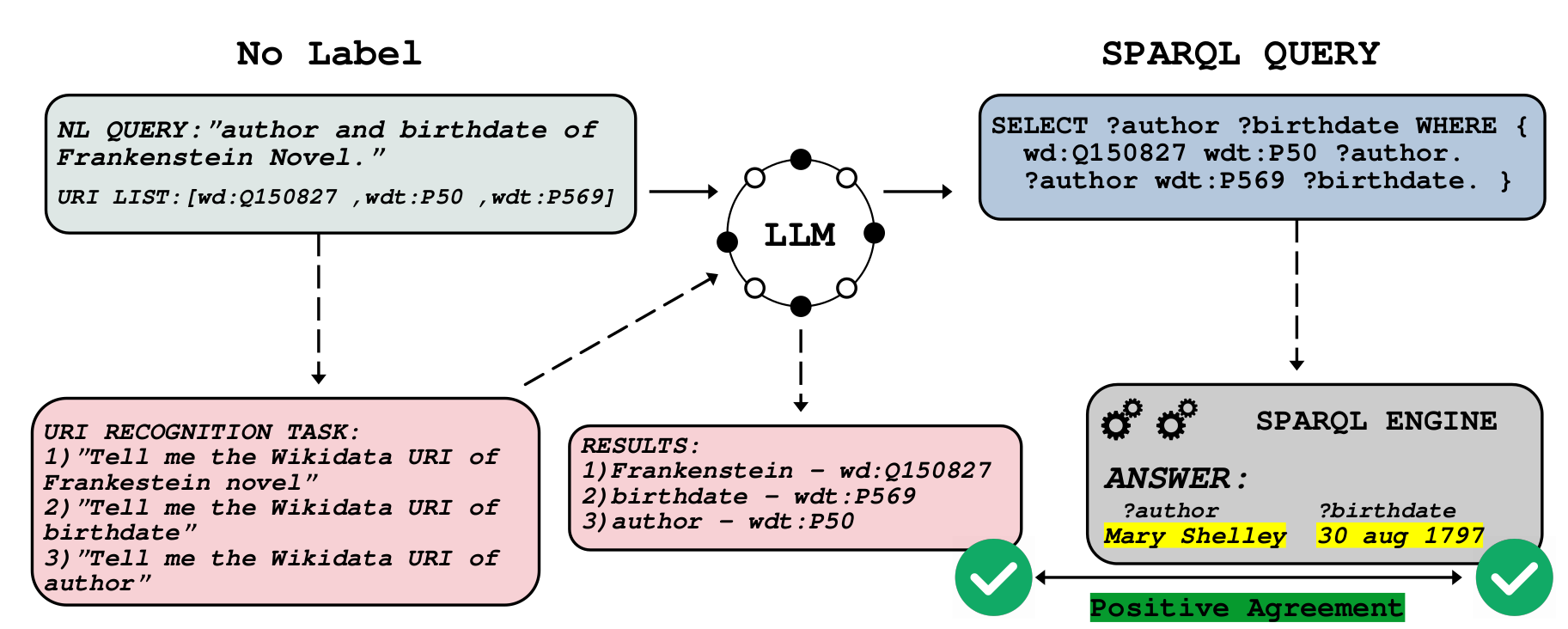}
    \caption{Lexical protoknowledge Impact analysis}
    \label{fig:lexical_protoknowledge_impact_image}
\end{figure}

\newpage

\subsection{\emph{Topological protoknowledge} analysis: Wikidata vs DBpedia}

\begin{figure}[h!]
    \centering
    \includegraphics[scale=0.24]{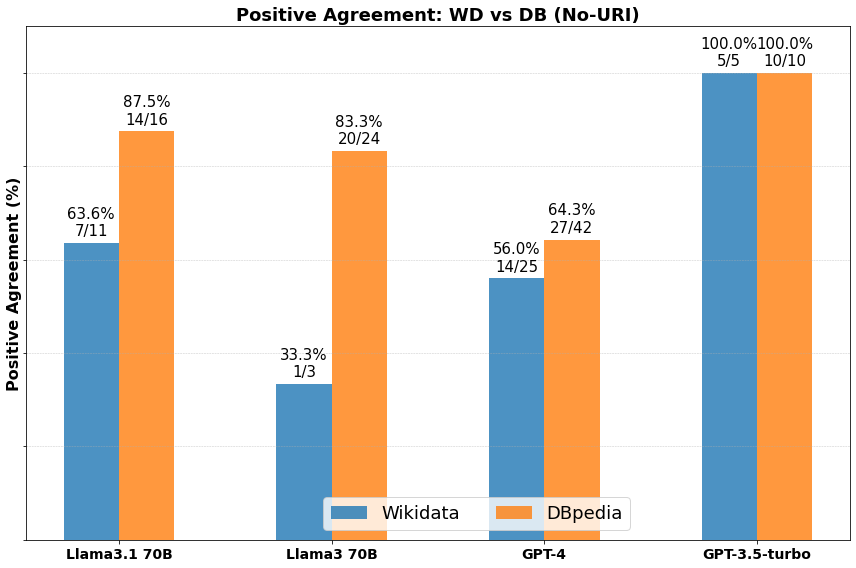}
    \caption{Wikidata vs DBpedia Positive Agreement in \texttt{No URI} setting.}
    \label{fig:no_uri_DC_correlation}
\end{figure}

\begin{figure}[h!]
    \centering
    \includegraphics[scale=0.24]{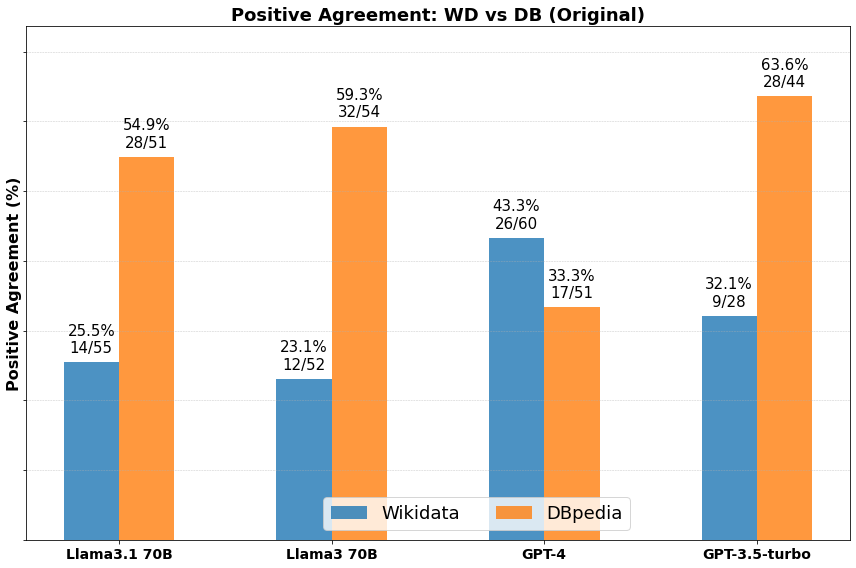}
    \caption{Wikidata vs DBpedia Positive Agreement in \texttt{Original} setting.}
    \label{fig:original_DC_correlation}
\end{figure}

\end{document}